\documentclass[runningheads]{llncs}
\usepackage[T1]{fontenc}
\usepackage{array}
\usepackage{graphicx}
\usepackage{mathalfa}
\usepackage{amsfonts,amssymb}
\usepackage{multirow}
\usepackage{amsmath}
\usepackage{marvosym}
\usepackage{ifsym}

\usepackage[colorlinks=true, linkcolor=blue, citecolor=blue, urlcolor=blue]{hyperref}

\newtheorem{myDef}{Definition}

\begin{document}
\title{M$^2$HGCL: Multi-Scale Meta-Path Integrated Heterogeneous Graph Contrastive Learning}

\titlerunning{M$^2$HGCL: Multi-Scale Meta-Path Integrated HGCL}
%

\author{Yuanyuan Guo\inst{1,2} \and
Yu Xia \inst{1,2} \and Rui Wang\inst{1,2} \and Rongcheng Duan \inst{3} \and Lu Li \inst{3} \and
Jiangmeng Li\inst{1,2}$^{(\textrm{\Letter})}$}
\authorrunning{Y.Guo et al.}
%

\institute{Institute of Software Chinese Academy of Sciences, Beijing, China \and University of Chinese Academy of Sciences, Beijing, China \\
\email{\{guoyuanyuan2022,jiangmeng2019\}@iscas.ac.cn} \and
Defence Industry Secrecy Examination and Certification Center, Beijing, China}

\maketitle              
\begin{abstract}
Inspired by the successful application of contrastive learning on graphs, researchers attempt to impose graph contrastive learning approaches on heterogeneous information networks. Orthogonal to homogeneous graphs, the types of nodes and edges in heterogeneous graphs are diverse so that specialized graph contrastive learning methods are required. Most existing methods for heterogeneous graph contrastive learning are implemented by transforming heterogeneous graphs into homogeneous graphs, which may lead to ramifications that the valuable information carried by non-target nodes is undermined thereby exacerbating the performance of contrastive learning models. Additionally, current heterogeneous graph contrastive learning methods are mainly based on initial meta-paths given by the dataset, yet according to our deep-going exploration, we derive empirical conclusions: only initial meta-paths cannot contain sufficiently discriminative information; and various types of meta-paths can effectively promote the performance of heterogeneous graph contrastive learning methods. To this end, we propose a new multi-scale meta-path integrated heterogeneous graph contrastive learning (M$^2$HGCL) model, which discards the conventional heterogeneity-homogeneity transformation and performs the graph contrastive learning in a joint manner. Specifically, we expand the meta-paths and jointly aggregate the direct neighbor information, the initial meta-path neighbor information and the expanded meta-path neighbor information to sufficiently capture discriminative information. A specific positive sampling strategy is further imposed to remedy the intrinsic deficiency of contrastive learning, i.e., the hard negative sample sampling issue. Through extensive experiments on three real-world datasets, we demonstrate that M$^2$HGCL outperforms the current state-of-the-art baseline models.

\end{abstract}

\section{Introduction}

HINs are complex networks containing multiple types of nodes and edges, which possess different attributes and features. Compared with traditional homogeneous information networks, HINs contain richer semantic and structural information, which further empowers HINs to reflect more sophisticated relationships in the real world, thereby having wider applications, such as recommendation systems,  social network analysis, and bioinformatics. In recent years, graph neural networks (GNNs) have achieved impressive success. The variants of GNNs, heterogeneous graph neural networks (HGNNs) \cite{tang2015pte,zhang2019heterogeneous,wang2019heterogeneous,fu2020magnn,hu2020heterogeneous}, have also attracted widespread attention. However, existing HGNNs require a mass of annotation information for downstream tasks, which is expensive to obtain in practice. To address this issue, many researchers attempt to leverage self-supervised learning to capture discriminative information from HINs \cite{hwang2020self,park2020unsupervised}.

Self-supervised learning is a training approach that allows models to learn from raw input data by utilizing the inherent structure or patterns in the data without explicit supervision. Contrastive methods, a type of self-supervised learning, aim to create effective representations by bringing semantically similar pairs together and pushing dissimilar pairs apart. Inspired by the success of contrastive learning in computer vision \cite{he2020momentum,chen2020simple}, many researchers have applied contrastive learning to homogeneous networks, called graph contrastive learning.  \cite{hwang2020self,ren2019heterogeneous,jiang2021contrastive} extend such a learning paradigm to HINs and propose heterogeneous graph contrastive learning (HGCL). However, in practice, HGCL suffers from two critical issues: 1) existing methods are restricted to the initial meta-paths given by the dataset; 2) when implementing graph contrastive learning by pre-transforming heterogeneous graphs into homogeneous graphs, conventional methods inevitably discard valuable information from non-target nodes.



\begin{figure}[t]

\begin{center}
\includegraphics[width=1\textwidth]{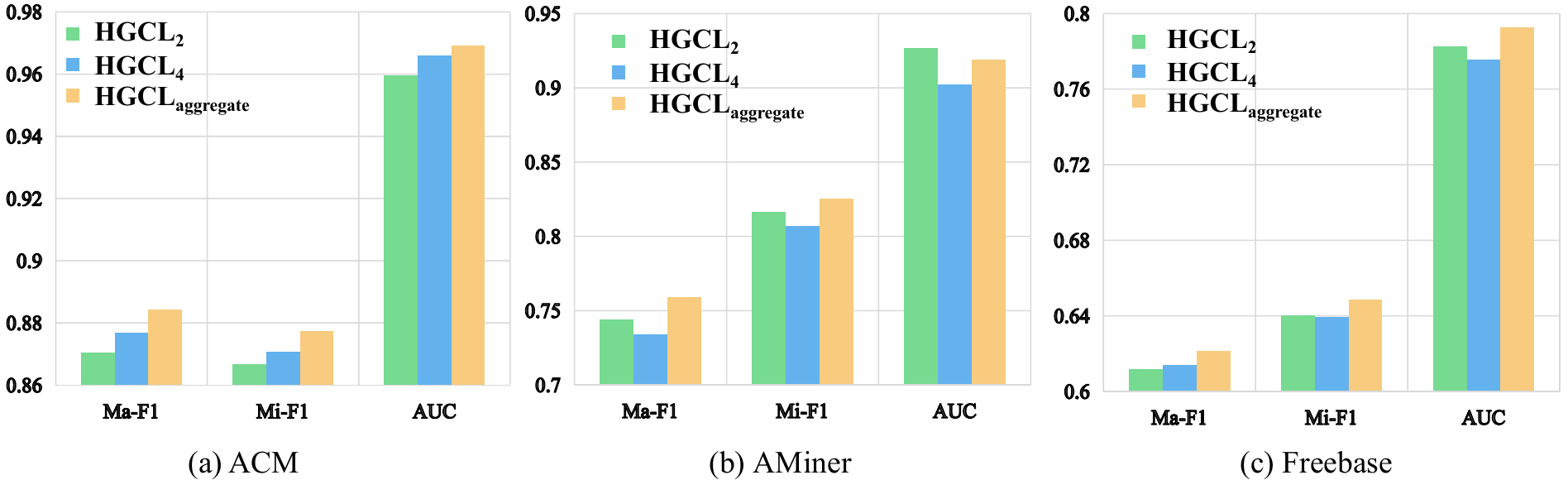}
\caption{Experiments to analyze the effects of meta-paths with different lengths on a specific HGCL model, i.e., HeCo \cite{wang2021self}. HGCL$_{\mathrm{2}}$ represents the model using initial 2-hop meta-paths. HGCL$_{\mathrm{4}}$ represents the model using expanded 4-hop meta-paths instead of the initial meta-paths. HGCL$_{\mathrm{aggregate}}$ denotes the model jointly using the initial and expanded meta-paths.}
\label{fig0}
\end{center}
\vspace{-2.0em}
\end{figure}

Regarding the first issue, we reckon that if HGCL is limited to short-chain information learned from the initial meta-paths, there is a problem of lacking global structural and contextual information. It may also fail to fully describe the semantic relationships between nodes, thus potentially unable to capture more complex semantic information. For example, in a movie recommendation system, if models only consider movies with the same director based on the initial 2-hop meta-paths, the other movies of the same genre may be unexpectedly ignored. Solely leveraging the long-chain information learned from the expanded meta-paths based on the initial meta-paths can improve the model to obtain global structural and contextual information, but such an approach may introduce noise and redundant information in the learned representations. For example, in a social network consisting of three types of nodes (people, companies, and schools) and two types of relationships (people-company/school), if an expanded 4-hop meta-path, e.g., people-company-people-company-people, is solely leveraged to learn the features of people, the learned features may contain certain redundant information because the third person node in this meta-path may have the same features as the second person node. Given the aforementioned issues, we propose a hypothesis that jointly aggregating the initial meta-path neighbors and the expanded meta-path neighbors may remedy the limitations of solely using one type of meta-paths, which further promotes the model's capacity to better capture the diverse semantic information of nodes. To verify our hypothesis, we design and conduct exploratory experiments on HeCo \cite{wang2021self}. The experimental results, as shown in Fig.~\ref{fig0}, demonstrate that the performance of the model aggregating the initial and expanded meta-paths exceed the performance of models solely using one type of meta-paths.

Considering the second issue, we provide the corresponding analyses that the conventional approach of directly transforming heterogeneous graphs into homogeneous graphs often involves a certain discriminative information loss. Different types of nodes and edges in HINs contain distinct semantic information, but after the transformation, only the information of relationships between nodes and edges of the same type is reserved, thereby resulting in the loss of specific discriminative information. For example, HGCML \cite{wang2022heterogeneous} transforms HINs into multiple subgraphs based on multiple meta-paths, ignoring the rich heterogeneous information contained by the non-target nodes. Despite utilizing heterogeneous information in HINs for contrastive learning, HeCo \cite{wang2021self} only aggregates the neighbor information of the target type in downstream tasks, disregarding the neighbor information of non-target types, thereby leading to information loss.


In light of the analyses of existing issues in the field of HGCL, we propose \textbf{M}ulti-scale \textbf{M}eta-path integrated \textbf{H}eterogeneous \textbf{G}raph \textbf{C}ontrastive \textbf{L}earning, dubbed M$^2$HGCL. Concretely, the contributions of this paper are three-fold:
\begin{itemize}
    \item[\textbullet] We introduce a novel multi-scale meta-path integrated heterogeneous graph contrastive learning approach, i.e., M$^2$HGCL, which conducts graph contrastive learning in a joint manner, without relying on the conventional heterogeneity-homogeneity transformation. 
    \item[\textbullet] To sufficiently exploit the diverse semantic relationships among different types of nodes and edges in HINs for heterogeneous graph contrastive learning, we propose a positive sampling strategy specific to HINs.
    \item[\textbullet] We perform comprehensive experiments on three widely-used HIN datasets to evaluate the effectiveness of our proposed model. The experimental results demonstrate that M$^2$HGCL outperforms the current state-of-the-art baselines, indicating the superiority of our approach for heterogeneous graph contrastive learning tasks.
\end{itemize}

\section{Related Work}
\subsection{Heterogeneous Graph Neural Networks}
Heterogeneous graph neural network (HGNN) has recently received much attention and there have been some models proposed. For example, HAN \cite{wang2019heterogeneous} argues that different types of edges should be assigned different weights and that different neighbor nodes should have distinct weights within the same type of edge. To achieve this, HAN employs node-level attention and semantic-level attention. Building on HAN, MAGNN \cite{fu2020magnn} generates node embeddings by applying node content transformation, meta-path aggregation, and inter-meta-path aggregation. To model large-scale heterogeneous graphs found on the web, scholars have proposed HGT \cite{hu2020heterogeneous}. HetGNN \cite{zhang2019heterogeneous}, operating in an unsupervised setting, leverages BiLSTM to capture both the heterogeneity of structure and content and is applicable to both transductive and inductive tasks. While the aforementioned methods have been successful, the majority of them are semi-supervised or supervised learning and depend heavily on the labeled data.
\subsection{Graph Contrastive Learning}
Graph contrastive learning(GCL) has emerged as one of the most critical techniques for unsupervised graph learning. Typically, a graph contrastive learning framework consists of a graph view generation component that generates positive and negative views, and a contrastive target that distinguishes positive pairs from negative pairs. DGI \cite{velickovic2019deep} achieves contrastive learning by maximizing the mutual information between the local representations and corresponding global representations of the graph. Besides, GraphCL \cite{you2020graph} employs augmentation techniques to create two views and then performs local-to-local contrast, while proposing four distinct augmentation methods. Moreover, to mitigate GraphCL's reliance on graph augmentation and negative samples, BGRL \cite{thakoor2021large} draws inspiration from BYOL \cite{grill2020bootstrap} and employs only basic graph augmentation techniques, thus eliminating the need for negative samples.
\subsection{Heterogeneous Graph Contrastive Learning}
Some researchers have extended graph contrastive learning to heterogeneous graphs and proposed the concept of heterogeneous graph contrastive learning. For instance, HDGI \cite{ren2019heterogeneous} applies DGI on heterogeneous graphs by leveraging meta-paths to model the semantic structure and by utilizing graph convolutional modules and semantic-level attention mechanisms to capture node representations. DMGI \cite{park2020unsupervised} introduces a contrastive learning approach based on single-view and multi-meta-path to enhance learning efficacy. It guides the fusion of different meta-paths through consensus regularization to capture complex relationships in graph data. HGCML \cite{wang2022heterogeneous} generates multiple subgraphs as multiple views using meta-paths and proposes a contrastive objective induced by meta-paths between views. HeCo \cite{wang2022heterogeneous} proposes to learn node embeddings by utilizing the network pattern and meta-path view of HIN to capture both local and high-order structures. However, while HGCML employs contrastive learning on heterogeneous graphs, it actually relies on contrastive learning on homogeneous graphs and does not fully utilize the rich heterogeneous information present in heterogeneous graphs. HeCo, on the other hand, employs heterogeneous information in contrastive learning training but does not do so in the final node embeddings.

\section{Preliminary}
\begin{figure}[t]
\begin{center}
\includegraphics[width=0.5\textwidth]{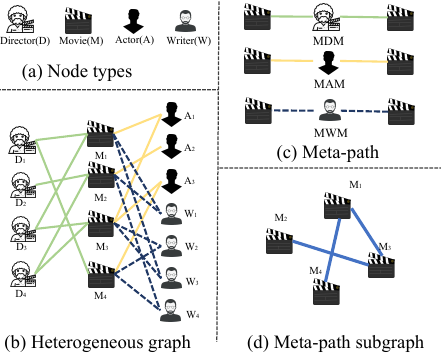}
\caption{An example of heterogeneous information networks (Freebase).} \label{fig1}
\end{center}
\vspace{-1.5em}
\end{figure}

In this section, we introduce the related concepts of heterogeneous graphs.
\begin{myDef}
\rm{\textbf{Heterogeneous Information Network(HIN).}}
\emph{HIN is defined as a graph network $G=(V, E, A, R)$, where $V$ denotes the set of nodes, $E$ denotes the set of edges, and $A$ and $R$ are the type sets of nodes and edges respectively, noting that $|A|+|R|>2$. Moreover, the set of nodes and edges $V, E$ are associated with a node-type mapping function $\phi: V\rightarrow A$ and an edge-type mapping function $\varphi: E\rightarrow R$. }
\end{myDef}

\begin{myDef}
\rm{\textbf{Meta-path.}} \emph{Meta-path $P_m$ is a path composed of multiple relations between nodes in a HIN, defined as $a_1\rightarrow a_2\rightarrow ...\rightarrow a_{n+1}$ (abbreviated as $a_1a_2a_{n+1}$), describing the compound relation between node $a_1$ and $a_{n+1}$. Note that $P_m\in P_M$, where $P_M$ is the set of meta-paths. For example, Fig.$\mathrm{~\ref{fig1}}$ (c) shows three meta-paths existing in the dataset Freebase: MDM represents two movies with the same director, MAM represents two movies with the same actor, and MWM represents two movies with the same writer.}
\end{myDef}

\begin{myDef}
\rm{\textbf{Meta-path Neighbors.}}
\emph{Given the meta-path $P_m$, the meta-path neighbors $N^{MP}_{P_m}$ is a set of nodes connected to the target node through the meta-path $P_m$. For example, in Fig.$\mathrm{~\ref{fig1}}$ (b), the meta-path neighbors of movie $M_3$ based on the meta-path MDM are movie $M_1$ and $M_2$.}
\end{myDef}

\begin{myDef}
\rm{\textbf{Meta-path Subgraph.}} 
\emph{Given the meta-path $P_m$, the subgraph $G^{P_m}$ represents the homogeneous graph including only the target node. Edges between any two nodes in $G^{P_m}$ depend on whether there exists meta-path $P_m$ between the corresponding nodes in the heterogeneous graph. For example, in Fig.$\mathrm{~\ref{fig1}}$, (d) represents the subgraph $G^{MDM}$ based on the meta-path MDM.}
\end{myDef}

\section{Methodology}

\begin{figure}[t]
\begin{center}
\includegraphics[width=1\textwidth]{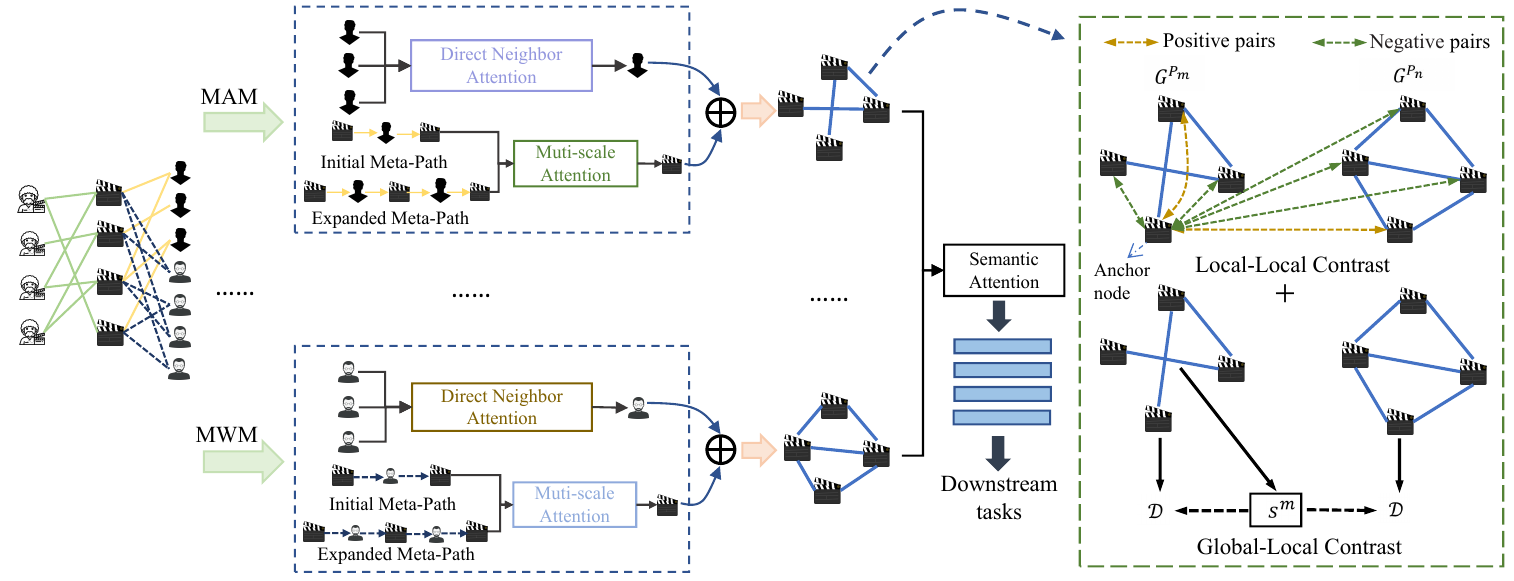}
\caption{The overall architecture of $\rm{M^2HGCL}$.} \label{fig2}
\end{center}
\vspace{-1.5em}
\end{figure}

In this section, we propose M$^2$HGCL, a novel heterogeneous graph contrastive learning method that integrates multi-scale meta-paths. The overall architecture of our model is shown in Fig.~\ref{fig2}. We first generate the subgraph based on every initial meta-paths and aggregate the direct neighbors, the initial meta-path neighbors, and the expanded meta-path neighbors of the target node as the node representation of this subgraph. Moreover, considering the specificity of HINs, we further design a positive sampling strategy for HINs.

\subsection{Node Feature Transformation}
Due to the existence of multiple types of nodes in HIN, each of which is in different feature spaces, they cannot be directly used for training. Therefore, we refer to HeCo \cite{wang2022heterogeneous} and firstly transform the features of all types of nodes into the same latent vector space. Specifically, for a node $i$ with type ${\phi_i}$, we use a matrix $\mathbf{W}_{\phi_i}$ specific to the node type to perform the following transformation on the initial feature $x_i$ of this node:
\begin{equation}
h_i = \sigma (\mathbf{W}_{\phi_i}\cdot x_i + \mathbf{b}_{\phi_i}) ,
\end{equation}
where $h_i$ is the initial node embedding after transformation, $\sigma$ is an activation function, and $\mathbf{b}_{\phi_i}$ denotes as vector bias.

\subsection{Muti-scale Meta-path Neighbors Aggregation}
In this subsection, we discard the conventional heterogeneity-homogeneity transformation, and instead propose a joint manner for each subgraph generated based on the initial meta-path that aggregates the direct neighbors, the initial meta-path neighbors, and the expanded meta-path neighbors to fully capture the discriminative information.

Since the meta-path is given in each subgraph, the node type of per target node's direct neighbors based on this meta-path is the same. For example, in Fig.~\ref{fig1}, the direct neighbors of the movie M$_3$ based on the meta-path MAM are actors A$_1$ and A$_3$, and they are both of the type `actor'. HAN \cite{wang2019heterogeneous} believes that same-type neighbor nodes do not have uniform contributions to the target node. In other words, it is important to consider the unique impact of each neighbor node on the target node, rather than treating them as completely identical or equivalent. Therefore, we adopt the attention mechanism to aggregate different neighbor node information to the target node. Specifically, we do the following aggregation:
\begin{equation}
h^{OH}_{i, P_m} = \sigma(\sum_{j\in N^{OH}_{P_m}} \zeta ^{OH}_{i,j} \cdot h_j),
\end{equation}
where $N^{OH}_{P_m}$ is the set of the direct one-hop neighbors of the target node $i$ based on meta-path $P_m$. $h_j$ is the transformed initial node embedding of the direct neighbor, $\sigma$ is the activation function, and $\zeta ^{OH}_{i,j}$ represents the attention value between the direct neighbor and the target node $i$, as shown below:
\begin{equation}
\zeta ^{OH}_{i,j} = \frac{e^{\mathrm{LeakyReLU}(\gamma ^{T}_{OH,P_m}\cdot[h_i||h_j])}}{\sum_{t\in N^{OH}_{P_m}}e^{\mathrm{LeakyReLU}(\gamma ^{T}_{OH,P_m}\cdot[h_i||h_t])}} ,
\end{equation}
where $\gamma _{OH, P_m}$ is the learnable parameter vector based on meta-path $P_m$. || is the concatenate operation.


Then, based on the given initial meta-path, we expand it to a 4-hop meta-path. For example, for the meta-path MAM, the expanded 4-hop meta-path is MAMAM. Next, we encode the subgraph based on initial and expanded meta-paths separately using GCN~\cite{kipf2016semi} encoder $\varepsilon$, as illustrated below:
\begin{equation}
\mathbf{H}^{I-P_m} = \varepsilon(\mathbf{H}, \mathbf{X}^{I-P_m}),
\end{equation}
\begin{equation}
\mathbf{H}^{E-P_m} = \varepsilon(\mathbf{H}, \mathbf{X}^{E-P_m}),
\end{equation}
where $\mathbf{H}$ is the initial node feature matrix, and $\mathbf{X}^{I-P_m}$ and $\mathbf{X}^{E-P_m}$ are the adjacency matrices of subgraphs generated based on the given initial meta-path and its expanded meta-path, respectively.

And we use muti-scale attention to aggregate the representations of nodes based on the initial and expanded meta-path, which can be expressed as follows:
\begin{equation}
 h^{AGG}_{i,P_m}= \omega_{I-P_m}\cdot h^{I-P_m}_i+\omega_{E-P_m}\cdot h^{E-P_m}_i ,
\end{equation}
here, $\omega_{I-P_m}$ and $\omega_{E-P_m}$ represent the importance weights of the initial meta-path and expanded meta-path, respectively. And $\omega_{I-P_m}+\omega_{E-P_m}=1$. Note that $h^{I-P_m}_i\in\mathbf{H}^{I-P_m}$ and $h^{E-P_m}_i\in\mathbf{H}^{E-P_m}$. Taking $\omega_{I-P_m}$ as an example, its computation is shown as follows:

\begin{equation} 
\qquad
\begin{aligned}
    \omega_{I-P_m} &=\frac{e^{\eta_{I-P_m}}}{e^{\eta_{I-P_m}}+e^{\eta_{E-P_m}}},  \\
    \eta_{I-P_m} &=\frac{1}{|V|}\sum_{i\in V} \mathrm{tanh}(\gamma ^{T}\cdot[\mathbf{W}_{MP}h^{I-P_m}_i+\mathbf{b}_{MP}]), 
\end{aligned}
\end{equation}
where $\mathbf{W}_{MP}$ and $\mathbf{b}_{MP}$ are the learnable parameters, and $\gamma$ represents the learnable shared attention vector.

Finally, the node representation of each subgraph consists of two parts: the direct neighbors and the aggregated representation of neighbors based on the initial meta-path and expanded meta-path. These two parts are concatenated to form the node representation of each subgraph. The method is as follows:
\begin{equation}
\mathbf{H}^{P_m} = [h^{OH}_{0,P_m}||h^{AGG}_{0,P_m}, h^{OH}_{1,P_m}||h^{AGG}_{1,P_m},...,h^{OH}_{|V_{P_m}|-1,P_m}||h^{AGG}_{|V_{P_m}|-1,P_m}],
\end{equation}
where $|V_{P_m}|$ denotes the number of nodes in meta-path subgraph $G^{P_m}$.

\subsection{Contrastive Learning with Positive Sampling Strategy}
To overcome the issue of hard negative samples in the classic GCL, we propose to select the most similar node as the positive sample from a semantic perspective. We take all subgraphs as positive and negative views in a pairwise manner. Then, we consider nodes that have meta-paths to the anchor node in the positive view and the nodes corresponding to the anchor node in the negative view as positive samples, while the remaining nodes are considered negative samples. Specifically, if there exists a meta-path $P_m$(limited to the initial meta-path) between the anchor node and $k$ other nodes in the positive subgraph $G^{P_m}$, then the number of positive samples for the anchor node is $k+1$, which is different from the conventional GCL method that only utilizes one positive sample per anchor node. We illustrate the significance of this approach with practical examples. For instance, in the ACM dataset, two articles written by the same author are likely to belong to the same field. Therefore, we consider these two articles as a positive sample pair during the contrastive learning process, aiming to reduce the distance between them and classify them as closely as possible.

Moreover, in order to learn rich semantic information from different meta-path subgraphs, we propose a new contrastive objective, which consists of two parts: one maximizes the local and global mutual information, and the other discriminates between local and local.

Firstly, we aim to maximize the mutual information between node representations (i.e., local representations) and the summary vector that captures the global information content of the entire graph. To achieve this goal, we define the local-global contrast objective as follows:
\begin{equation}
\mathcal{L}^{m,n}_{global}(i)=-log(\mathcal{D}(h^m_i,s^m))-log(\mathcal{D}(h^n_i,s^m)),
\end{equation}
where $s^m$ is the summary vector obtained from the subgraph 
$G^{P_m}$ through the READOUT($\cdot$) function (mean pooling in this paper). $h^m_i$ and $h^n_i$ denote the representation of node $i$ in the meta-path subgraph $G^{P_m}$ and $G^{P_n}$, and $\mathcal{D}(h^m_i,s^m) = \sigma(h^m_i\mathbf{W}s^m)$. $\mathbf{W}$ is a learnable matrix. 

Secondly, the local-local contrast objective aims to learn discriminative node representations in order to enhance the usability of node representations in downstream tasks. Specifically, based on the proposed positive sampling strategy, the local-local contrast objective is defined as follows:
\begin{equation}
\qquad
\begin{aligned}
    \mathcal{L}^{m,n}_{local}(i, \mathbb{P}_i) &=-log
\frac{pos\_sim}{pos\_sim+neg\_sim},  \\
    pos\_sim &=\sum_{v\in\mathbb{P}_i}e^{sim(h^m_i,h_v)/\tau}, \\
    neg\_sim &= \sum_{v\notin\mathbb{P}_i,v\in V_{P_m}}e^{sim(h^m_i,h_v)/\tau}+\sum_{v\notin\mathbb{P}_i,v\in V_{P_n}}e^{sim(h^n_i,h_v)/\tau},
\end{aligned}
\end{equation}
 where $\tau$ denotes the temperature parameter that controls the distribution of the data. $h_i^m$ and $h_i^n$ are the representation of node $i$ in the meta-path subgraph $G^{P_m}$ and $G^{P_n}$. $\mathbb{P}_i$ represents the set of positive samples corresponding to node $i$ based on the proposed positive sampling strategy. $V_{P_m}$ and $V_{P_n}$ are the sets of nodes in the $G^{P_m}$ and $G^{P_n}$.

Finally, the overall objective:
\begin{equation}
\label{eqloss}
\mathcal{J}=\sum_{m\in {P_M}}\sum_{n\in {P_M}}\sum_{i\in V}(\alpha\cdot\mathcal{L}^{m,n}_{global}(i)+(1-\alpha)\cdot\mathcal{L}^{m,n}_{local}(i, \mathbb{P}_i)),
\end{equation}
where $P_M$ is the set of meta-paths. $\alpha$ is a weight coefficient and $\alpha\in[0,1]$. $m,n$ denote $G^{P_m}$ and $G^{P_n}$. Ultimately, we use a semantic-level attention mechanism to integrate all subgraphs generated based on the initial meta-paths, producing the final representation of the target node for downstream tasks.

\section{Experiment}
\subsection{Experimental Setup}
\subsubsection{Datasets} To demonstrate the superiority of our model over state-of-the-art methods, we conduct extensive experiments on three HIN datasets. Dataset statistics and details are summarized in Table \ref{tab2}.

\begin{table}[t]
\renewcommand\arraystretch{1.0}
\setlength{\belowcaptionskip}{-0.15cm} 
\caption{The statistics of the datasets.}\label{tab2}
\begin{center}
\begin{tabular}{|c|c|c|c|c|}
\hline
Dataset  & Node                                                                                                      & Meta-path                                               & Target node & Num of class \\ \hline
AMiner \cite{hu2019adversarial}   & \begin{tabular}[c]{@{}c@{}}paper(P):6,564\\ author(A):13,329\\ reference(R):35,890\end{tabular}               & \begin{tabular}[c]{@{}c@{}}PAP\\ PRP\end{tabular}       & paper       &4              \\ \hline
ACM \cite{zhao2020network}      & \begin{tabular}[c]{@{}c@{}}paper(P):4,019\\ author(A):7,167\\ subject(S):60\end{tabular}                     & \begin{tabular}[c]{@{}c@{}}PAP\\ PSP\end{tabular}       & paper       & 3            \\ \hline
Freebase \cite{li2021leveraging} & \begin{tabular}[c]{@{}c@{}}movie(M):3,492\\ actor(A):33,401\\ director(D):2,502\\ writer(W):4,459\end{tabular} & \begin{tabular}[c]{@{}c@{}}MAM\\ MDM\\ MWM\end{tabular} & movie       &3              \\ \hline
\end{tabular}
\end{center}
\vspace{-1.5em}
\end{table}

\subsubsection{Baselines} We compare $\mathrm{M^2HGCL}$ with ten baselines, including six unsupervised heterogeneous methods, one semi-supervised heterogeneous method, and three unsupervised homogeneous methods. The unsupervised heterogeneous methods are Mp2vec \cite{dong2017metapath2vec}, HERec  \cite{shi2018heterogeneous}, HetGNN \cite{zhang2019heterogeneous}, DMGI \cite{park2020unsupervised}, HGCML \cite{wang2022heterogeneous} and HeCo \cite{wang2021self}. The semi-supervised method is HAN \cite{wang2019heterogeneous}. The unsupervised homogeneous methods are GraphSAGE \cite{hamilton2017inductive}, GAE \cite{kipf2016variational}, and DGI \cite{velickovic2019deep}.

\subsubsection{Implementation Details} The expanded meta-path denotes the 4-hop meta-path obtained by expanding the initial 2-hop meta-path. For optimizing our model, we utilize Glorot initialization to initialize our model parameters and employ Adam as the optimization algorithm during training. For the temperature parameter $\tau$ and $\alpha$ in the overall objective, we test ranging from 0.1 to 0.9 with step 0.1 to select the optimal value. For the number of hidden layer nodes, AMiner and Freebase are 64, and ACM is 128. Besides, we determine the learning rate for our model, setting it at 3e-3 for the AMiner dataset, 5e-4 for the ACM dataset, and 1e-3 for the Freebase dataset. In meta-path neighbor encoding, we leverage 1-layer GCN as the encoder. For a fair comparison, we randomly run the experiments 5 times and report the average results with standard deviations. For HGCML, we conduct experiments with 40\% and 60\% labeled nodes, and report the results by following the official implementation. HeCo and other competitors' results are reported from \cite{wang2021self}.

\subsection{Node Classification} 

\label{nodeclass}
\begin{table}[t]
\setlength{\belowcaptionskip}{0.2cm} 
\caption{Quantitative results on node classification with 40\% labeled nodes. The best results are marked in bold, and the second-best are underlined.}\label{tab3}
\renewcommand\arraystretch{1.0}
\begin{tabular}{|c|c|ccc|ccc|ccc|}
\hline
\multirow{2}{*}{Methods(40\%)}   & \multirow{2}{*}{Data}  & \multicolumn{3}{c|}{AMiner}                                                     & \multicolumn{3}{c|}{ACM}                                                        & \multicolumn{3}{c|}{Freebase}                                                   \\ \cline{3-11} 
                           &                        & Ma-F1                    & Mi-F1                    & AUC                       & Ma-F1                    & Mi-F1                    & AUC                       & Ma-F1                    & Mi-F1                    & AUC                       \\ \hline
\multirow{2}{*}{HAN}       & \multirow{2}{*}{X,A,Y} & 63.85                    & 76.89                    & 80.72                     & 87.47                    & 87.21                    & 94.84                     & 59.63                    & 63.74                    & 77.74                     \\
                           &                        & \multicolumn{1}{l}{±1.5} & \multicolumn{1}{l}{±1.6} & \multicolumn{1}{l|}{±2.1} & \multicolumn{1}{l}{±1.1} & \multicolumn{1}{l}{±1.2} & \multicolumn{1}{l|}{±0.9} & \multicolumn{1}{l}{±2.3} & \multicolumn{1}{l}{±2.7} & \multicolumn{1}{l|}{±1.2} \\ \hline
\multirow{2}{*}{GraphSAGE} & \multirow{2}{*}{X,A}   & 45.77                    & 52.10                    & 74.44                     & 55.96                    & 60.98                    & 71.06                     & 44.88                    & 57.08                    & 66.42                     \\
                           &                        & \multicolumn{1}{l}{±1.5} & \multicolumn{1}{l}{±2.2} & \multicolumn{1}{l|}{±1.3} & \multicolumn{1}{l}{±6.8} & \multicolumn{1}{l}{±3.5} & \multicolumn{1}{l|}{±5.2} & \multicolumn{1}{l}{±4.1} & \multicolumn{1}{l}{±3.2} & \multicolumn{1}{l|}{±4.7} \\ \hline
\multirow{2}{*}{GAE}       & \multirow{2}{*}{X,A}   & 65.66                    & 71.34                    & 88.29                     & 61.61                    & 66.38                    & 79.14                     & 52.44                    & 56.05                    & 74.05                     \\
                           &                        & \multicolumn{1}{l}{±1.5} & \multicolumn{1}{l}{±1.8} & \multicolumn{1}{l|}{±1.0} & \multicolumn{1}{l}{±3.2} & \multicolumn{1}{l}{±1.9} & \multicolumn{1}{l|}{±2.5} & \multicolumn{1}{l}{±2.3} & \multicolumn{1}{l}{±2.0} & \multicolumn{1}{l|}{±0.9} \\ \hline
\multirow{2}{*}{Mp2vec}    & \multirow{2}{*}{X,A}   & 64.77                    & 69.66                    & 88.82                     & 62.41                    & 64.43                    & 80.48                     & 57.80                    & 61.01                    & 75.51                     \\
                           &                        & \multicolumn{1}{l}{±0.5} & \multicolumn{1}{l}{±0.6} & \multicolumn{1}{l|}{±0.2} & \multicolumn{1}{l}{±0.6} & \multicolumn{1}{l}{±0.6} & \multicolumn{1}{l|}{±0.4} & \multicolumn{1}{l}{±1.1} & \multicolumn{1}{l}{±1.3} & \multicolumn{1}{l|}{±0.8} \\ \hline
\multirow{2}{*}{HERec}     & \multirow{2}{*}{X,A}   & 64.50                    & 71.57                    & 88.70                     & 61.21                    & 62.62                    & 79.84                     & 59.28                    & 62.71                    & 76.08                     \\
                           &                        & \multicolumn{1}{l}{±0.7} & \multicolumn{1}{l}{±0.7} & \multicolumn{1}{l|}{±0.4} & \multicolumn{1}{l}{±0.8} & \multicolumn{1}{l}{±0.9} & \multicolumn{1}{l|}{±0.5} & \multicolumn{1}{l}{±0.6} & \multicolumn{1}{l}{±0.7} & \multicolumn{1}{l|}{±0.4} \\ \hline
\multirow{2}{*}{HetGNN}    & \multirow{2}{*}{X,A}   & 58.97                    & 68.47                    & 83.14                     & 72.02                    & 74.46                    & 85.01                     & 48.57                    & 53.96                    & 69.48                     \\
                           &                        & \multicolumn{1}{l}{±0.9} & \multicolumn{1}{l}{±2.2} & \multicolumn{1}{l|}{±1.6} & \multicolumn{1}{l}{±0.4} & \multicolumn{1}{l}{±0.8} & \multicolumn{1}{l|}{±0.6} & \multicolumn{1}{l}{±0.5} & \multicolumn{1}{l}{±1.1} & \multicolumn{1}{l|}{±0.2} \\ \hline
\multirow{2}{*}{DGI}       & \multirow{2}{*}{X,A}   & 54.72                    & 63.87                    & 77.86                     & 80.23                    & 80.41                    & 91.52                     & 53.40                    & 57.82                    & 72.97                     \\
                           &                        & \multicolumn{1}{l}{±2.6} & \multicolumn{1}{l}{±2.9} & \multicolumn{1}{l|}{±2.1} & \multicolumn{1}{l}{±3.3} & \multicolumn{1}{l}{±3.0} & \multicolumn{1}{l|}{±2.3} & \multicolumn{1}{l}{±1.4} & \multicolumn{1}{l}{±0.8} & \multicolumn{1}{l|}{±1.1} \\ \hline
\multirow{2}{*}{DMGI}      & \multirow{2}{*}{X,A}   & 61.92                    & 63.60                    & 88.02                     & 86.23                    & 86.02                    & 96.35                     & 49.88                    & 54.28                    & 70.77                     \\
                           &                        & \multicolumn{1}{l}{±2.1} & \multicolumn{1}{l}{±2.5} & \multicolumn{1}{l|}{±1.3} & \multicolumn{1}{l}{±0.8} & \multicolumn{1}{l}{±0.9} & \multicolumn{1}{l|}{±0.3} & \multicolumn{1}{l}{±1.9} & \multicolumn{1}{l}{±1.6} & \multicolumn{1}{l|}{±1.6} \\ \hline
\multirow{2}{*}{HGCML}      & \multirow{2}{*}{X,A}   &\underline{74.13}              &\underline{85.11}              &\underline{92.87}               &\underline{90.36}              & \textbf{90.32}           &96.29               &\underline{62.57}              &\underline{70.49}              &\underline{79.12}               \\
                           &                        & \multicolumn{1}{l}{\underline{±0.6}} & \multicolumn{1}{l}{\underline{±0.2}} & \multicolumn{1}{l|}{\underline{±0.1}} & \multicolumn{1}{l}{\underline{±0.1}} & \multicolumn{1}{l}{\textbf{±0.3}} & \multicolumn{1}{l|}{±0.2} & \multicolumn{1}{l}{\underline{±0.1}} & \multicolumn{1}{l}{\underline{±0.2}} & \multicolumn{1}{l|}{\underline{±0.2}} \\ \hline
\multirow{2}{*}{HeCo}      & \multirow{2}{*}{X,A}   &73.75              &80.53              &92.11               &87.61              & 87.45           &\underline{96.40}               &61.19              &64.03             &78.44             \\
                           &                        & \multicolumn{1}{l}{±0.5} & \multicolumn{1}{l}{±0.7} & \multicolumn{1}{l|}{±0.6} & \multicolumn{1}{l}{±0.5} & \multicolumn{1}{l}{±0.5} & \multicolumn{1}{l|}{\underline{±0.4}} & \multicolumn{1}{l}{±0.6} & \multicolumn{1}{l}{±0.7} & \multicolumn{1}{l|}{±0.5} \\ \hline
\multirow{2}{*}{$\mathrm{M^2HGCL}$}    & \multirow{2}{*}{X,A}   & \textbf{76.04}           & \textbf{86.79}           & \textbf{94.65}            & \textbf{91.48}           & \underline{90.10}              & \textbf{97.19}            & \textbf{64.04}           & \textbf{72.50}           & \textbf{81.09}            \\
                           &                        & \multicolumn{1}{l}{\textbf{±0.2}}    & \multicolumn{1}{l}{\textbf{±0.6}}    & \multicolumn{1}{l|}{\textbf{±0.5}}    & \multicolumn{1}{l}{\textbf{±0.2}} & \multicolumn{1}{l}{\underline{±0.2}} & \multicolumn{1}{l|}{\textbf{±0.3}} & \multicolumn{1}{l}{\textbf{±0.1}} & \multicolumn{1}{l}{\textbf{±0.4}}   & \multicolumn{1}{l|}{\textbf{±0.3}} \\ \hline
\end{tabular}
\vspace{-0.5em}
\end{table}

\begin{table}[t]
\setlength{\belowcaptionskip}{0.2cm}
\caption{Quantitative results on node classification with 60\% labeled nodes. The best results are marked in bold, and the second-best are underlined.}\label{60}
\renewcommand\arraystretch{1.0}
\begin{tabular}{|c|c|ccc|ccc|ccc|}
\hline
\multirow{2}{*}{Methods(60\%)}   & \multirow{2}{*}{Data}  & \multicolumn{3}{c|}{AMiner}                                                     & 
\multicolumn{3}{c|}{ACM}                                                        & \multicolumn{3}{c|}{Freebase}                                                   \\ \cline{3-11} 
                           &                        & Ma-F1                    & Mi-F1                    & AUC                       & Ma-F1                    & Mi-F1                    & AUC                       & Ma-F1                    & Mi-F1                    & AUC                       \\ \hline
\multirow{2}{*}{HAN}       & \multirow{2}{*}{X,A,Y} & 62.02                    & 74.73                    & 80.39                     & 88.41                    & 88.10                    & 94.68                     & 56.77                    & 61.06                    & 75.69                     \\
                           &                        & \multicolumn{1}{l}{±1.2} & \multicolumn{1}{l}{1.4}  & \multicolumn{1}{l|}{±1.5} & \multicolumn{1}{l}{±1.1} & \multicolumn{1}{l}{±1.2} & \multicolumn{1}{l|}{±1.4} & \multicolumn{1}{l}{±1.7} & \multicolumn{1}{l}{±2.0} & \multicolumn{1}{l|}{±1.5} \\ \hline
\multirow{2}{*}{GraphSAGE} & \multirow{2}{*}{X,A}   & 44.91                    & 51.36                    & 74.16                     & 56.59                    & 60.72                    & 70.45                     & 45.16                    & 55.92                    & 66.78                     \\
                           &                        & \multicolumn{1}{l}{±2.0} & \multicolumn{1}{l}{±2.2} & \multicolumn{1}{l|}{±1.3} & \multicolumn{1}{l}{±5.7} & \multicolumn{1}{l}{±4.3} & \multicolumn{1}{l|}{±6.2} & \multicolumn{1}{l}{±3.1} & \multicolumn{1}{l}{±3.2} & \multicolumn{1}{l|}{±2.0} \\ \hline
\multirow{2}{*}{GAE}       & \multirow{2}{*}{X,A}   & 63.74                    & 67.70                    & 86.92                     & 61.67                    & 65.71                    & 77.90                     & 50.65                    & 53.85                    & 71.75                     \\
                           &                        & \multicolumn{1}{l}{±1.6} & \multicolumn{1}{l}{±1.9} & \multicolumn{1}{l|}{±0.8} & \multicolumn{1}{l}{±2.9} & \multicolumn{1}{l}{±2.2} & \multicolumn{1}{l|}{±2.8} & \multicolumn{1}{l}{±0.4} & \multicolumn{1}{l}{±0.4} & \multicolumn{1}{l|}{±0.4} \\ \hline
\multirow{2}{*}{Mp2vec}    & \multirow{2}{*}{X,A}   & 60.65                    & 63.92                    & 85.57                     & 61.13                    & 62.72                    & 79.33                     & 55.94                    & 58.74                    & 74.78                     \\
                           &                        & \multicolumn{1}{l}{±0.3} & \multicolumn{1}{l}{±0.5} & \multicolumn{1}{l|}{±0.2} & \multicolumn{1}{l}{±0.4} & \multicolumn{1}{l}{±0.3} & \multicolumn{1}{l|}{±0.4} & \multicolumn{1}{l}{±0.7} & \multicolumn{1}{l}{±0.8} & \multicolumn{1}{l|}{±0.4} \\ \hline
\multirow{2}{*}{HERec}     & \multirow{2}{*}{X,A}   & 65.53                    & 69.76                    & 87.74                     & 64.35                    & 65.15                    & 81.64                     & 56.50                    & 58.57                    & 74.89                     \\
                           &                        & \multicolumn{1}{l}{±0.7} & \multicolumn{1}{l}{±0.8} & \multicolumn{1}{l|}{±0.5} & \multicolumn{1}{l}{±0.8} & \multicolumn{1}{l}{±0.9} & \multicolumn{1}{l|}{±0.7} & \multicolumn{1}{l}{±0.4} & \multicolumn{1}{l}{±0.5} & \multicolumn{1}{l|}{±0.4} \\ \hline
\multirow{2}{*}{HetGNN}    & \multirow{2}{*}{X,A}   & 57.34                    & 65.61                    & 84.77                     & 74.33                    & 76.08                    & 87.64                     & 52.37                    & 56.84                    & 71.01                     \\
                           &                        & \multicolumn{1}{l}{±1.1} & \multicolumn{1}{l}{±2.2} & \multicolumn{1}{l|}{±0.9} & \multicolumn{1}{l}{±0.6} & \multicolumn{1}{l}{±0.7} & \multicolumn{1}{l|}{±0.7} & \multicolumn{1}{l}{±0.8} & \multicolumn{1}{l}{±0.7} & \multicolumn{1}{l|}{±0.5} \\ \hline
\multirow{2}{*}{DGI}       & \multirow{2}{*}{X,A}   & 55.45                    & 63.10                    & 77.21                     & 80.03                    & 80.15                    & 91.41                     & 53.81                    & 57.96                    & 73.32                     \\
                           &                        & \multicolumn{1}{l}{±2.4} & \multicolumn{1}{l}{±3.0} & \multicolumn{1}{l|}{±1.4} & \multicolumn{1}{l}{±3.3} & \multicolumn{1}{l}{±3.2} & \multicolumn{1}{l|}{±1.9} & \multicolumn{1}{l}{±1.1} & \multicolumn{1}{l}{±0.7} & \multicolumn{1}{l|}{±0.9} \\ \hline
\multirow{2}{*}{DMGI}      & \multirow{2}{*}{X,A}   & 61.15                    & 62.51                    & 86.20                     & 87.97                    & 87.82                    & 96.79                     & 52.10                    & 56.69                    & 73.17                     \\
                           &                        & \multicolumn{1}{l}{±2.5} & \multicolumn{1}{l}{±2.6} & \multicolumn{1}{l|}{±1.7} & \multicolumn{1}{l}{±0.4} & \multicolumn{1}{l}{±0.5} & \multicolumn{1}{l|}{±0.2} & \multicolumn{1}{l}{±0.7} & \multicolumn{1}{l}{±1.2} & \multicolumn{1}{l|}{±1.4} \\ \hline
\multirow{2}{*}{HGCML}      & \multirow{2}{*}{X,A}   &\underline{76.26}              &\underline{86.10}              &\underline{93.41}               &\underline{91.02}              & \underline{91.10}          &\underline{97.49}               &\underline{65.17}              &\underline{71.08}              &\underline{79.24}               \\
                           &                        & \multicolumn{1}{l}{\underline{±0.7}} & \multicolumn{1}{l}{\underline{±0.8}} & \multicolumn{1}{l|}{\underline{±0.6}} & \multicolumn{1}{l}{\underline{±0.2}} & \multicolumn{1}{l}{\underline{±0.4}}
                           & \multicolumn{1}{l|}{\underline{±0.4}} & \multicolumn{1}{l}{\underline{±0.3}} & \multicolumn{1}{l}{\underline{±0.1}} & \multicolumn{1}{l|}{\underline{±0.1}} \\ \hline
\multirow{2}{*}{HeCo}      & \multirow{2}{*}{X,A}   & 75.80              & 82.46              & 92.40               & 89.04              & 88.71           & 96.55               & 60.13             & 63.61              & 78.04              \\
                           &                        & \multicolumn{1}{l}{±1.8} & \multicolumn{1}{l}{±1.4} & \multicolumn{1}{l|}{±0.7} & \multicolumn{1}{l}{±0.5} & \multicolumn{1}{l}{±0.5} & \multicolumn{1}{l|}{±0.3} & \multicolumn{1}{l}{±1.3} & \multicolumn{1}{l}{±1.6} & \multicolumn{1}{l|}{±0.4} \\ \hline
\multirow{2}{*}{$\mathrm{M^2HGCL}$}    & \multirow{2}{*}{X,A}   & \textbf{77.56}           & \textbf{87.17}           & \textbf{94.88}            & \textbf{91.41}           & \textbf{91.29}              & \textbf{98.43}            & \textbf{66.61}           & \textbf{73.02}           & \textbf{81.96}            \\
                           &                        & \multicolumn{1}{l}{\textbf{±0.7}} & \multicolumn{1}{l}{\textbf{±0.6}} & \multicolumn{1}{l|}{\textbf{±0.3}} & \multicolumn{1}{l}{\textbf{±0.1}} & \multicolumn{1}{l}{\textbf{±0.2}} & \multicolumn{1}{l|}{\textbf{±0.3}} & \multicolumn{1}{l}{\textbf{±0.4}} & \multicolumn{1}{l}{\textbf{±0.2}} & \multicolumn{1}{l|}{\textbf{±0.2}} \\ \hline
\end{tabular}

\end{table}

We utilize the learned node embeddings to train a linear classifier for evaluating our model. We randomly select 40\% and 60\% of labeled nodes from each dataset for training, and reserve 1,000 nodes for validation and another 1,000 nodes for testing. Our evaluation metric for node classification on the test set is Macro-F1, Micro-F1, and AUC, which are reported as the average results with standard deviations. The results are shown in Table \ref{tab3} and \ref{60}. It is evident that $\mathrm{M^2HGCL}$ generally achieves the best performance among all baseline methods and all splits, even outperforming semi-supervised learning methods, i.e., HAN. This indicates the effectiveness of fully utilizing heterogeneous information in HIN and the importance of expanded meta-paths. However, we consider that the ACM dataset's `subject' nodes may have had a relatively smaller representation compared to other node types, resulting in only marginal differences in the node embeddings. This could be a contributing factor to our model's underperformance on the ACM dataset when classifying. Notably, the AMiner dataset suffers from label imbalance, with the number of objects in the label with the maximum number of nodes being 7 times greater than that in the label with the minimum number of nodes. This demonstrates the effectiveness of the proposed model in practical scenarios with label imbalance.

\subsection{Node Clustering}
\begin{table}[t]
\begin{center}
\caption{Quantitative results on node clustering. The best results are marked in bold, and the second-best results are underlined.}\label{tab4}
\renewcommand\arraystretch{1.0}
\begin{tabular}{|c|cc|cc|cc|}
\hline
Datasets  & \multicolumn{2}{c|}{AMiner}     & \multicolumn{2}{c|}{ACM}        & \multicolumn{2}{c|}{Freebase} \\ \hline
Metrics   & NMI            & ARI            & NMI            & ARI            & NMI           & ARI           \\ \hline
GraphSage & 15.74          & 10.10          & 29.20          & 27.72          & 9.05          & 10.49         \\ \hline
GAE       & 28.58          & 20.90          & 27.42          & 24.49          & 19.03         & 14.10         \\ \hline
Mp2vec    & 30.80          & 25.26          & 48.43          & 34.65          & 16.47         & 17.32         \\ \hline
HERec     & 27.82          & 20.16          & 47.54          & 35.67          & 19.76         & 19.36         \\ \hline
HetGNN    & 21.46          & 26.60          & 41.53          & 34.81          & 12.25         & 15.01         \\ \hline
DGI       & 22.06          & 15.93          & 51.73          & 41.16          & 18.34         & 11.29         \\ \hline
DMGI      & 19.06          & 20.09          & 51.66          & 46.64          & 16.98         & 16.91         \\ \hline
HGCML     & \underline{36.10}         & \underline{35.29}        & \textbf{65.13} & \textbf{62.77} & 15.46         & 14.29         \\ \hline
HeCo      & 32.26          & 28.64          & 56.87 & 56.94 & \underline{20.38}         & \underline{20.98}         \\ \hline
$\mathrm{M^2HGCL}$     & \textbf{39.14} & \textbf{37.00} & \underline{59.19}          & \underline{58.11}          & \textbf{20.96}             &\textbf{21.23}             \\ \hline
\end{tabular}
\end{center}
\vspace{-1.5em}
\end{table}

\begin{table}[t]
\setlength{\belowcaptionskip}{-0.2cm} 
\caption{Ablation study of $\mathrm{M^2HGCL}$ for pretext tasks on node classification.}\label{tab5}
\begin{center}
\renewcommand\arraystretch{1.0}
\begin{tabular}{|c|ccc|ccc|ccc|}
\hline
Variants & \multicolumn{3}{c|}{AMiner}                      & \multicolumn{3}{c|}{ACM}                         & \multicolumn{3}{c|}{Freebase}                    \\ \hline
$\mathrm{M^2HGCL}_{w/o\ expanded}$        & 73.12          & 79.21          & 93.17          & 88.74          & 88.54          & 96.92          & 61.33          & 64.21          & 79.34          \\ \hline
$ \mathrm{M^2HGCL}_{w/o\ direct}$          & 73.14          & 78.30          & 93.37          & 85.91          & 85.53          & 94.82          & 59.94          & 63.19          & 78.34          \\ \hline
$ \mathrm{M^2HGCL}_{w/o\ global}$          & 74.26          & 80.06          & 94.03          & 76.22          & 76.04          & 88.77          & 61.57          & 64.00          & 80.46          \\ \hline
$ \mathrm{M^2HGCL}_{w/o\ local}$          & 71.89          & 79.02          & 93.23          & 86.19          & 85.88          & 95.05          & 56.23          & 58.17          & 77.11          \\ \hline
$ \mathrm{M^2HGCL}_{w/o\ p-samp}$          & 74.73          & 80.93          & 93.64          & 80.47          & 81.17          & 89.95          & 58.68          & 61.29          & 78.22          \\ \hline
$\mathrm{M^2HGCL}$         & \textbf{77.56}     & \textbf{87.17}           & \textbf{94.88}     & \textbf{91.41}           & \textbf{91.29}    
& \textbf{98.43}           & \textbf{66.61}      & \textbf{73.02}           & \textbf{81.96} \\ \hline
\end{tabular}
\end{center}
\vspace{-0.8em}
\end{table}

We further perform K-means clustering to verify the quality of learned node embeddings. We use two widely-used clustering evaluation metrics, namely normalized mutual information (NMI) and adjusted rand index (ARI), to compare the performance of our proposed model with other baseline methods. The results are presented in Table \ref{tab4}. As we can see, our proposed model achieves significant improvements over other baseline methods on the AMiner and Freebase datasets. Specifically, on the Freebase dataset, our model outperforms approximately 36\% on NMI and 49\% on ARI compared to HGCML, demonstrating the superiority and effectiveness of our approach. However, the results on the ACM dataset fall short of other baseline methods. As analyzed in Section \ref{nodeclass}, the relatively smaller number of `subject' nodes in the ACM dataset compared to other node types, which may have resulted in only marginal differences in the final node representations, is the primary reason for our underperformance on the ACM dataset when clustering.


\subsection{Ablation Study}

\begin{figure}[t]
\begin{center}
\includegraphics[width=1.0\textwidth]{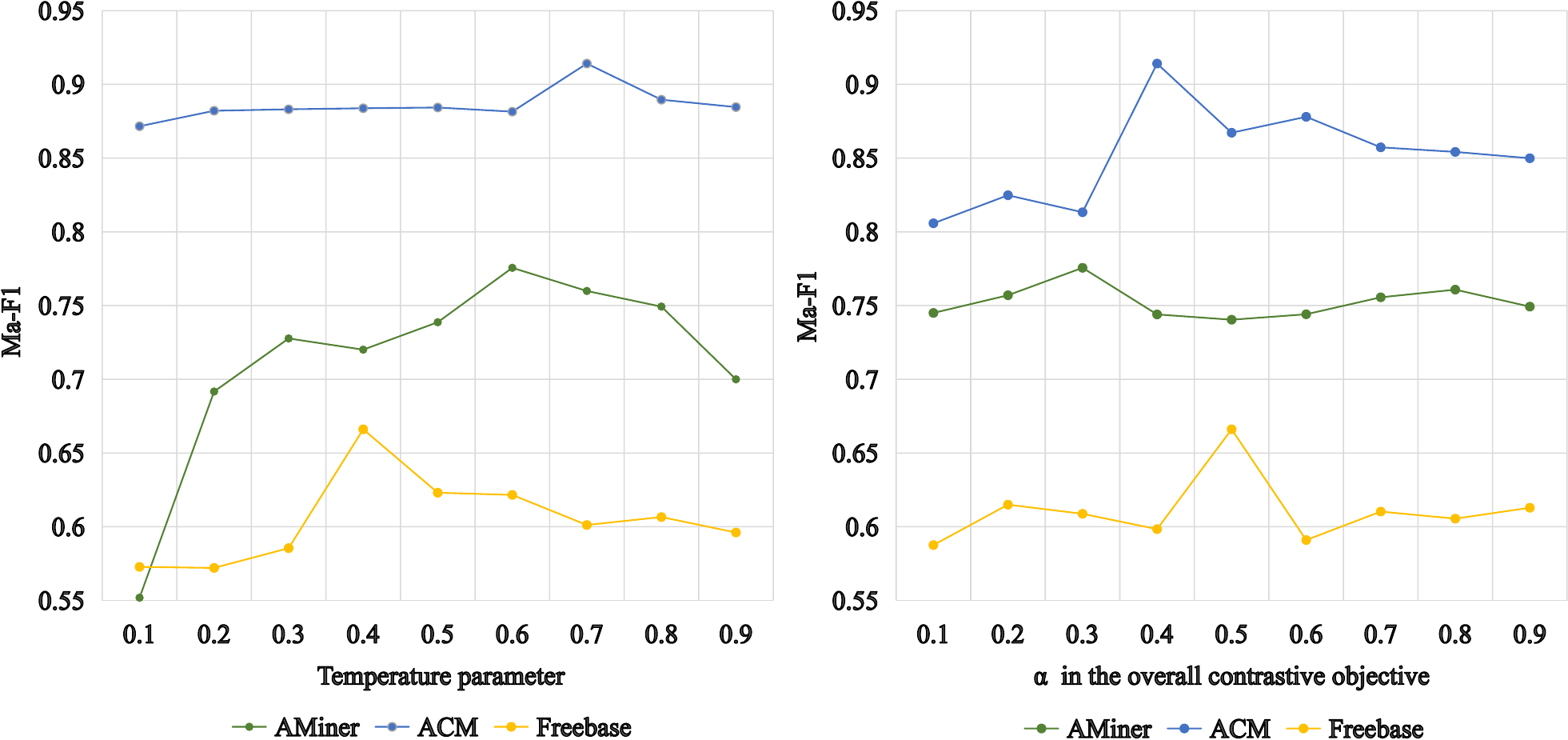}
\caption{Analysis of the temperature parameter $\tau$ and the weight coefficient $\alpha$ in the overall contrastive objective.} 
\label{fig3}
\end{center}
\vspace{-1.0em}
\end{figure}
We conduct the ablation study on our model, analyzing the effectiveness of its constituent components. To demonstrate the effectiveness of incorporating expanded meta-path neighbor information, we limit our model to the initial meta-path, referring to this variant as $\mathrm{M^2HGCL}_{w/o\ expanded}$. To underscore the importance of incorporating the direct neighbor information in final embeddings, we eliminate the direct neighbor information and solely employ meta-path neighbor information for the embedding of the subgraph, denoting this variant as $\mathrm{M^2HGCL}_{w/o\ direct}$. Additionally, we individually remove the local-global and local-local contrast objective to evince the indispensability of these two components in the contrastive objective and dub these two variants as $\mathrm{M^2HGCL}_{w/o\ global}$ and $ \mathrm{M^2HGCL}_{w/o\ local}$. Finally, to highlight the efficacy of our positive sampling strategy in enhancing model performance, we substitute our proposed strategy with the traditional GCL positive sample definition method, calling this variant $\mathrm{M^2HGCL}_{w/o\ p-samp}$. We present the experimental results of all variants in Table \ref{tab5}, and the results evince that $\mathrm{M^2HGCL}$ outperforms all other variants, thereby substantiating the efficacy and indispensability of each component of our proposed model.

\subsection{Analysis of Hyperparameter}

In this section, we analyze two crucial hyperparameters, the weight coefficient $\alpha$ in the overall contrastive objective and the temperature parameter $\tau$, that play a crucial role in model training. To explore their sensitivity, we conduct a comprehensive evaluation of the model's performance on node classification.

Fig.~\ref{fig3} demonstrates the model's performance on three datasets for different values of $\tau$ and $\alpha$.  From Fig.~\ref{fig3}, we can observe that as  $\tau$ and $\alpha$ increase, the performance of the model exhibits an initial rise in volatility, followed by a decrease across all three datasets, among which the ACM dataset exhibits a smaller variation range with increasing $\tau$, but there still exists a peak performance point. Specifically, the optimal values for the AMiner dataset are $\tau$=0.6 and $\alpha$=0.3, for the ACM dataset are $\tau$=0.7 and $\alpha$=0.4, and for the Freebase dataset are $\tau$=0.4 and $\alpha$=0.5. $\tau$ is observed to be less than or equal to 0.5 across all datasets, indicating the dominant impact of global-local loss over local-local loss based on Equation~\ref{eqloss}. This insight provides valuable information on the contribution of different loss terms to our model performance.

\section{Conclusion}
We propose a novel model for heterogeneous graph contrastive learning that integrates multi-scale meta-paths to address existing issues in HGCL. Our model represents target nodes using direct neighbor information, the initial neighbor information, and the expanded meta-paths neighbor information. We introduce a positive sampling strategy for HINs and weight the local-local and local-global contrast objectives to enhance model performance. Experiments show that our proposed model outperforms other state-of-the-art methods.


%
%
%
\bibliographystyle{splncs04}
\bibliography{ref}

\end{document}